\documentclass{article}
\usepackage{spconf,amsmath,graphicx}
\usepackage{booktabs}
\usepackage{amssymb}
\usepackage{color}
\usepackage{soul}
\usepackage{multirow}
\usepackage{hyperref}
\usepackage{tikz}

\newcommand\copyrighttext{%
  \footnotesize \textcopyright 2022 IEEE. Personal use of this material is permitted.
  Permission from IEEE must be obtained for all other uses, in any current or future
  media, including reprinting/republishing this material for advertising or promotional
  purposes, creating new collective works, for resale or redistribution to servers or
  lists, or reuse of any copyrighted component of this work in other works. }

    \newcommand\mycopyrightnotice{%
\begin{tikzpicture}[remember picture,overlay]
\node[anchor=south,yshift=10pt] at (current page.south) {\fbox{\parbox{\dimexpr\textwidth-\fboxsep-\fboxrule\relax}{\copyrighttext}}};
\end{tikzpicture}%
}

\title{Improving Generalization of Deep Networks for Estimating Physical Properties of Containers and Fillings}

\name{Hengyi Wang$^{1*}$, Chaoran Zhu$^{2}$\sthanks{Equally contributed.}, Ziyin Ma$^3$, Changjae Oh$^2$ }
\address{$^1$ School of Computer Science, University College London \\
$^2$ School of Electronic Engineering and Computer Science, Queen Mary University of London \\
$^3$ School of Computer Science, University of Birmingham}

\begin{document}
\ninept

\maketitle

\begin{abstract}
We present methods to estimate the physical properties of household containers and their fillings manipulated by humans. We use a lightweight, pre-trained convolutional neural network with coordinate attention as a backbone model of the pipelines to accurately locate the object of interest and estimate the physical properties in the CORSMAL Containers Manipulation (CCM) dataset. We address the filling type classification with audio data and then combine this information from audio with video modalities to address the filling level classification. For the container capacity, dimension, and mass estimation, we present a data augmentation and consistency measurement to alleviate the over-fitting issue in the CCM dataset caused by the limited number of containers. We augment the training data using an object-of-interest-based re-scaling that increases the variety of physical values of the containers. We then perform the consistency measurement to choose a model with low prediction variance in the same containers under different scenes, which ensures the generalization ability of the model. Our method improves the generalization ability of the models to estimate the property of the containers that were not previously seen in the training.
\vspace{-2pt}

\end{abstract}
\mycopyrightnotice
\begin{keywords}
Multi-modal perception, generalization, data augmentation
\vspace{-7pt}
\end{keywords}

\section{Introduction}
\vspace{-2pt}
\label{sec:intro}
Human-robot collaboration is an important concept in the modern industry, where robots are required to interact and cooperate with humans~\cite{sanchez2020benchmark,medina2016human,rosenberger2020object,ortenzi2021object,yang2020reactive,pang2021towards}. In this work, we aim at addressing tasks that are crucial in human-robot object handover: estimating the physical properties of target containers and their fillings. A good estimation model should generalize to containers with various materials, shapes, textures, and transparency. However, estimating the physical properties of the container is still challenging due to the invisible content, occlusion, and variety of household containers.

Existing methods~\cite{liu2021va2mass,ishikawa2021audio,iashin2021top} address the classification of filling level and type mostly with audio data, whereas the estimation of capacity, dimension, and mass of the container are remaining problems. A dataset available and commonly used to estimate these properties is the CORSMAL Containers Manipulation (CCM) dataset~\cite{corsmal_dataset}. Training a neural network for regressing the capacity, dimension and mass value on this dataset, which with only 9 objects appears repeatedly in the total 684 videos, can be easily over-fitted and eventually make the regression similar to classification. Therefore, neural networks trained directly on the dataset, such as~\cite{iashin2021top}, usually lead to poor generalization. Other works~\cite{liu2021va2mass,ishikawa2021audio} that apply non-deep-learning approaches for estimating the container capacity show certain generalization ability but have limited performance.

In this paper, we present methods to estimate the physical properties of household containers, such as filling type and level classification, container capacity, mass, and dimension estimation using the CCM dataset~\cite{corsmal_dataset}.
We use MobileNetv2~\cite{sandler2018mobilenetv2} with the coordinate attention (CA)~\cite{hou2021coordinate} as a backbone model, with minor changes of input modalities for each task. Our backbone model incorporates positional information, which shows good generalization ability with both audio and RGB-Depth (RGBD) inputs. The audio data is converted to clips of log-Mel spectrogram~\cite{logan2000mel} which are used as input for the filling type classification. To address the filling level classification, we jointly combine the intermediate audio feature from the filling type classification with the intermediate visual feature from the corresponding RGB frames. For the estimation of container capacity, dimensions and mass, we use RGBD images cropped by YOLO-v5~\cite{yolov5} to filter out the irrelevant information in the target frame. To address the generalization problem, we augment the annotations for container capacity, dimension, and mass using the geometric relationship between 2D images and 3D objects by randomly resizing the detected object of interest and changing the value of the label based on the geometric relationship.
We further present a variance-based evaluation that measures the consistency of the model in each container to evaluate the generalization ability of trained models.

\begin{figure}[t]
    \centering
    \vspace{10pt}
    \includegraphics[width=1.0\columnwidth]{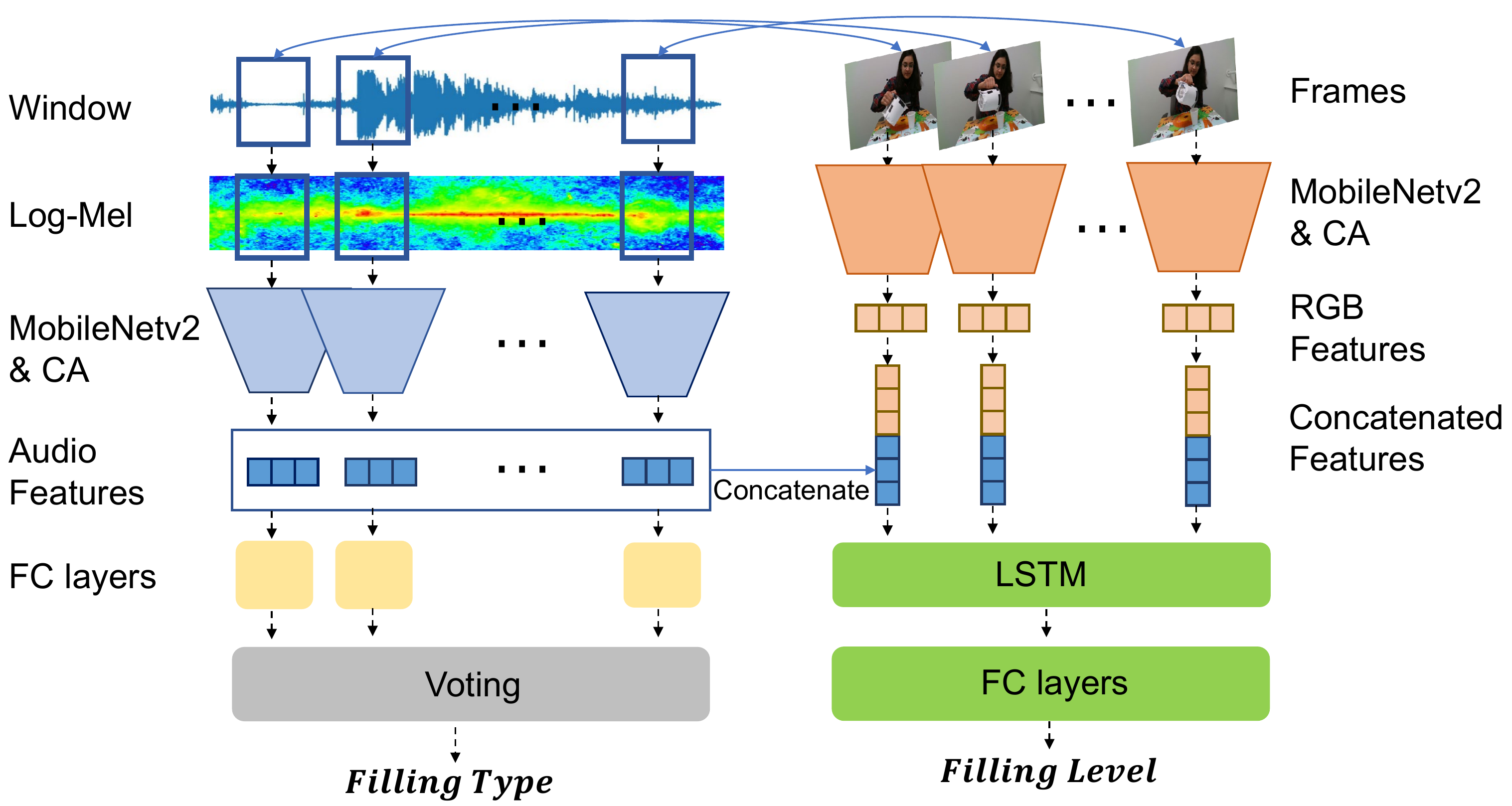}
    \caption{Overview of our approach for the filling level and type classification of the container. Key-- CA: Coordinate Attention, FC: Fully Connected, LSTM: Long Short Term Memory, Log-Mel: Log-Mel spectrogram.}
    \label{Fig:fillingtype}
\end{figure}

\begin{figure*}[t]
    \centering
    \includegraphics[width=0.9\textwidth]{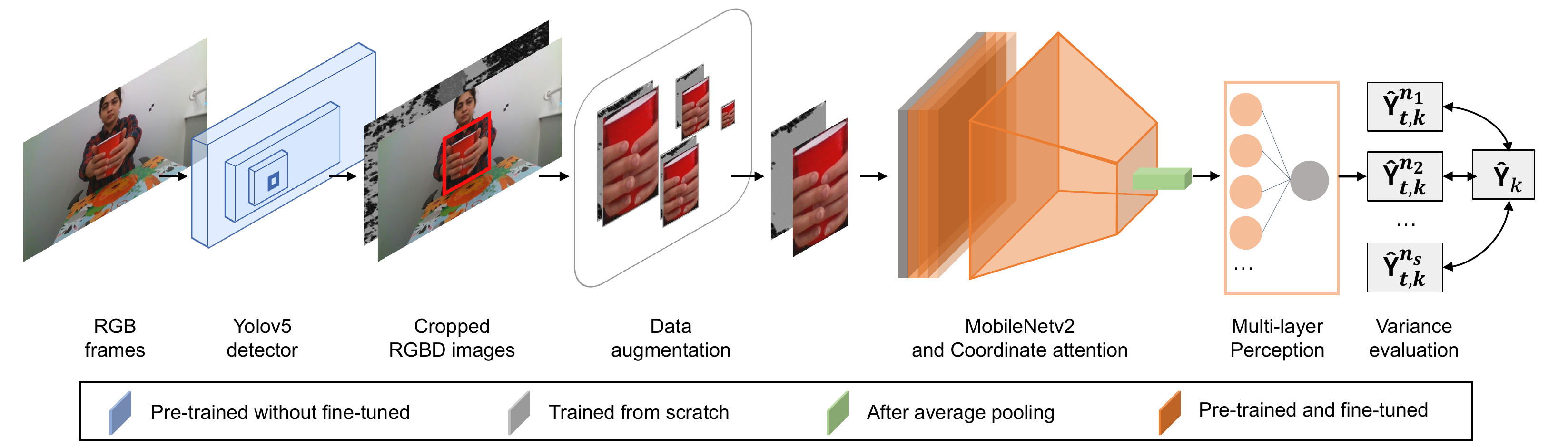}
    \caption{Overview of our framework to estimate container capacity, mass, and dimension. We use the same network architecture for these tasks and employ the data augmentation strategy to augment the target objects detected from YOLO-v5~\cite{yolov5}. At training time, we perform training with RGBD input and evaluate the variance of the videos with the same container $k$. }
    \label{fig:Overview}

\end{figure*}

\vspace{-2pt}
\section{Proposed Method}
We present our methods to perform filling level and type classification (Section 2.1) and container capacity, mass, and dimension estimation (Section 2.2).
We use the CCM dataset~\cite{corsmal_dataset} that provides audio-visual recordings where people manipulate containers including cups, drinking glass, and food boxes by pouring the filling into the container, shaking the food box, or performing no action. The container filling is labeled with one of four classes, $\{ \text{\textit{empty}}, \text{\textit{pasta}}, \text{\textit{rice}}, \text{\textit{water}} \}$, with three different filling levels, $\{ 0\%, 50\%, 90\% \}$. Among the video data from four views, we use the RGB and depth images captured from a camera facing the human operator.

\subsection{Filling level and type classification}

Figure~\ref{Fig:fillingtype} shows the pipeline of estimating the filling type and level of the container. Inspired by~\cite{ishikawa2021audio}, we also rely on the clipped audios and voting strategy for the filling type estimation and we adopt the LSTM~\cite{hochreiter1997long} network passed by the intermediate feature from the filling type estimation to predict the filling level. However, instead of using the VGG backbone \cite{vgg} on the audio features, we use a more efficient backbone, MobileNetv2~\cite{sandler2018mobilenetv2} with CA~\cite{hou2021coordinate}. For the filling level classification, we employ a multi-modal approach that integrates information from audio and RGB video. We combine the intermediate audio feature from the filling type classification with the RGB feature since combining different modalities can utilize more relevant object features and improve the generalization of the model in a small training set.

\noindent
\textbf{Backbone model.} To emphasize the information for the target tasks, we include the CA~\cite{hou2021coordinate} to MobileNetv2~\cite{sandler2018mobilenetv2}. Unlike the Squeeze-and-Excitation attention~\cite{hu2018squeeze} which neglects the positional information, the CA generates the channel attention with the spatial information, which captures the useful information from the audio feature and RGB frames by encoding the long-range positional information.

\noindent
\textbf{Feature extraction.} An audio sequence usually contains various events, which may not be relevant to our objective. Training a model on such audio sequences may introduce noise caused by other events. Using additional existing annotation containing the starting and ending time of events in the audio signals ~\cite{ishikawa2021audio}, we divide the signals into a series of audio clips overlapping in time. The clips which do not contain the target events will be assigned to an empty label. These audio clips will be used for extracting the log-Mel spectrogram~\cite{logan2000mel} and the training will be performed on these clips instead of the whole audio sequence. The corresponding RGB video frames of each audio clip are also extracted to jointly estimate the filling level in a time-series manner.

\noindent
\textbf{Filling type classification.} Let us denote an $n$-th audio sequence with multiple clips, where the $t$-th audio feature map, i.e. the log-Mel spectrogram, $X_{t}^{n}$, is extracted using the audio clips with the length of 0.025s and the stride of 0.01s, with an assigned label, $Y_{t}^{n}$. In this task, we aim to find a model $f_{\theta_T}$, with learned parameters $\theta_T$ on the audio features, to perform the classification on the four classes of filling types. The objective function is:
\begin{equation}
	\theta_T=\mathop{\arg\min}_{\theta}{\sum\limits_{n}\sum\limits_{t}{\mathcal{L}}\left(f_{\theta}\left(X_{t}^{n}\right), Y_{t}^{n}\right)},
\label{fillingtype}
\vspace{-5pt}
\end{equation}
where $\mathcal{L}$ is the cross-entropy loss. At the inference time, we perform classification on each $X_{t}^{n}$ in audio. As each $X_{t}^{n}$ generates an independent prediction, we adopt a voting strategy to decide the final predictions. If the number of non-empty predictions does not exceed the threshold, which we empirically set to 5, the sequence is labelled to \textit{empty}. Otherwise, the label is the class with the most predictions.

\noindent
\textbf{Filling level classification.}
We adopt a multi-modal fusion strategy, which integrates information from multiple modalities. As each modality captures specific characteristics of the underlying information, combining multiple complementary modalities can provide a highly comprehensive set of data.
As the filling level of the containers varies over time, we perform this task in a time-series manner. We first align the frames with the center of the generated audio clips. These audio and RGB frames are then used as input to the pre-trained audio model ($f_{\theta_T}$) from the filling type estimation and the pre-trained vision model from ImageNet classification~\cite{deng2009imagenet}, respectively, since the CCM training data is not sufficient to train the whole pipeline from scratch. Note that the parameters for the pre-trained audio and vision models are fixed during training to avoid the model over-fit to the limited training samples.
We retrieve the embeddings before the fully connected (FC) layer from audio and vision models. The intermediate features from these two models are then concatenated and fed into a LSTM network~\cite{hochreiter1997long} with FC layers to classify the filling level.
We aim to find the model $g_{\theta_L}$ with the learned parameters $\theta_L$ as follows:

\vspace*{-0.7\baselineskip}
\begin{equation}
	\theta_L=\mathop{\arg\min}_{\theta}{\sum\limits_{n}{\mathcal{L}}(g_{\theta}(f_{\theta_T}\left(X_{1...t}^{n}), Z_{1...t}^{n}), Y^{n}\right)}
\label{fillingtype}
\end{equation}

\noindent
where $X_{1...t}^{n}, Z_{1...t}^{n}$ are the audio and RGB data, respectively.

\subsection{Container capacity, mass, and dimension estimation}
Figure~\ref{fig:Overview} shows the pipeline of estimating the capacity, mass, and dimensions. We utilize the same network structures {with previous tasks, i.e. the MobileNetv2 with CA, as the backbone model.} We first crop the object of interest using bounding boxes containing the target object. We then perform the data augmentation and use the augmented data as input to train the model for better generalization. The transfer learning strategy, which uses the trained parameters on the dimension estimation, is also adopted to improve the generalization ability of our model on capacity and mass estimation. After generating the predictions, we apply the proposed consistency evaluation to find the model with less variance, which may have a better generalization on those unseen containers.

\noindent
\textbf{Backbone model.} As an RGBD image requires 4 channel inputs, we initialize the parameters for the first $3$ input channels with MobileNetv2 pre-trained on ImageNet classification~\cite{deng2009imagenet} and randomly initialize the parameters for the $4$-th channel of our model. We also include the CA~\cite{hou2021coordinate} for better generalization of the model.

\noindent
\textbf{Container detection.}
Prior to estimating the capacity, mass, and dimension, we first locate the container's position in the images extracted from videos since the background does not contribute to the predictions directly but occupies a large portion in the image. We apply a YOLO-v5~\cite{yolov5} object detector to localize  the object in the RGB frame and crop the object of interest in both the RGB frame and corresponding depth image (already spatially aligned). As there is no annotation of bounding boxes in the CCM dataset, we use the detector pre-trained on COCO dataset~\cite{lin2014microsoft}.

\noindent
\textbf{Data augmentation.} Even if the training set of CCM contains 684 audio-visual recordings with different manipulations and conditions, the labels for container capacity, mass and dimensions are limited to the number of containers used in the set, i.e. 9. Considering that the model estimates the continuous values, training the model to address such a regression problem on the CCM dataset may cause an over-fitting issue. Assuming that, given a fixed depth of the target container, the capacity, dimension, and mass can be proportional to the number of pixels that belong to the target object, we make use of this relationship to perform the data augmentation. Specifically, we randomly resize the target container image by a scale factor $s$ and change their corresponding labels with such a scale factor. Considering that the depth value has not been changed and by neglecting the correction for the perspective projection of the object, we multiply $s^2$ to the capacity and mass and $s$ to the dimension values according to scaling the target container image by $s$. Note that this augmentation is an approximation that might not be physically correct. To compensate for the noise introduced by the proposed approach, we sample $s$ according to a normal distribution with the mean $\mu=1$ and the standard deviation $\sigma \in [0.5, 1.0]$. The distribution is truncated to make sure $s \in [0.5, 1.5]$. The final images are padded to $[640, 640]$ and resized to $[320, 320]$ for reducing the computation cost.

\begin{figure}[t]
    \centering
    \includegraphics[width=0.65\columnwidth]{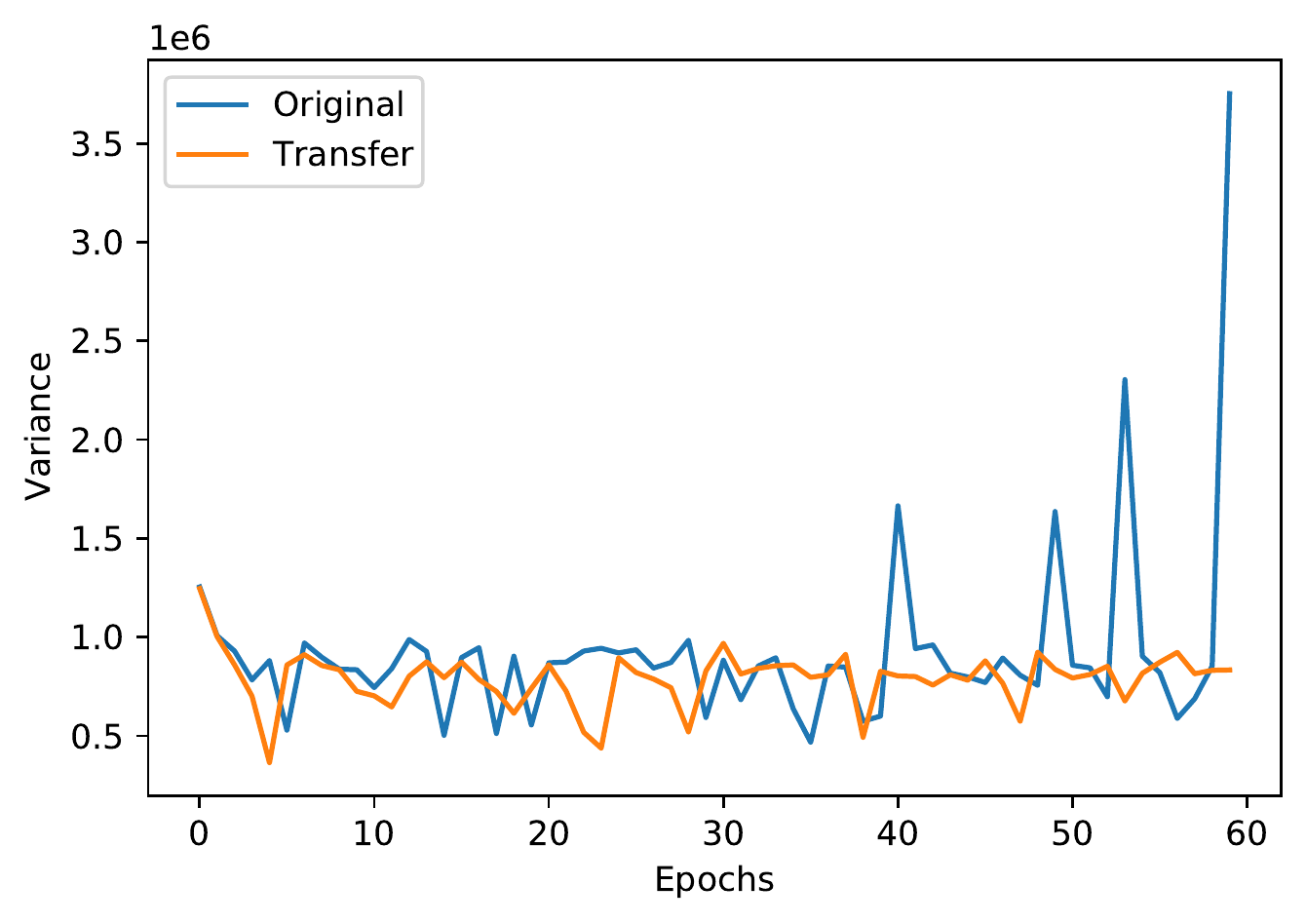}
    \vspace{-9pt}
    \caption{Effect of the transfer learning on capacity estimation.}
    \label{Fig:transfer}
    \vspace{-9pt}
\end{figure}

\noindent
\textbf{Transfer learning.} The container capacity and mass can be obtained if the dimensions and the materials (i.e density) are known. We thus consider that the container capacity and mass estimations are higher-level tasks than the dimension estimation as the model needs to not only detect the object shape but also learn semantic context, such as material and volume from the image. We thus employ a transfer learning that uses knowledge learned from the container dimension estimation for the container capacity and mass estimation. We first train our model on the dimension estimation and then use the pre-trained weights as initial weights to train models for the capacity and mass estimation. Figure~\ref{Fig:transfer} visualizes the intra-class variance in our validation dataset. We can observe that the model trained with the weights transferred from the dimension data can maintain a more stable variance curve compared to the model trained from scratch. The model can encode the representations of the target container after learning to predict the width and height, which improves the generalization ability of our model.

\begin{figure}[t]
    \centering
    \includegraphics[width=0.6\columnwidth]{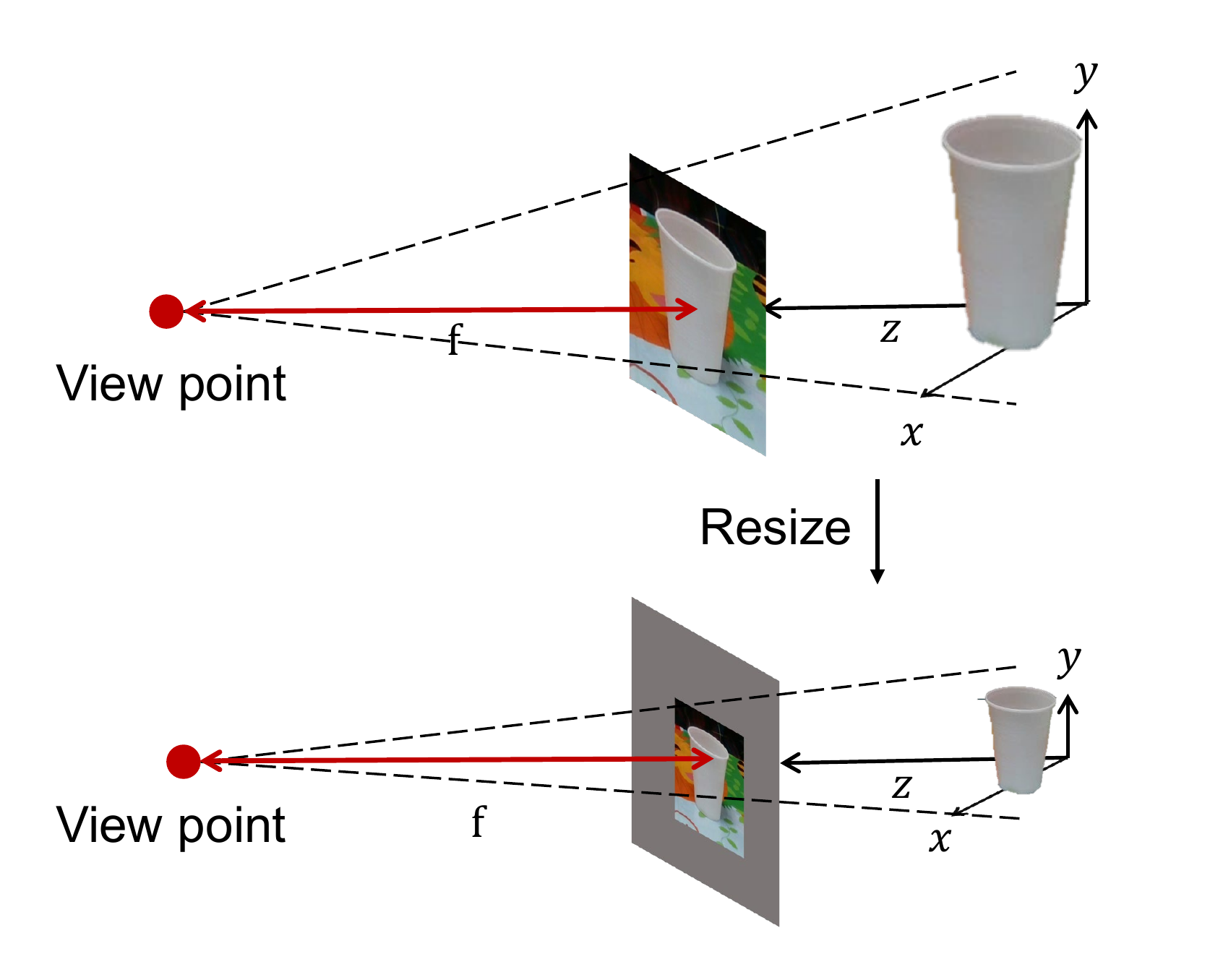}
    \vspace{-9pt}
    \caption{Illustration of our data augmentation. The object is resized while keeping the camera matrix and distance fixed. The image is resized to a fixed aspect ratio with zero padding, the gray area surrounding the container. {{$f$} is the focal length from the camera matrix, $x$,$y$,$z$ are the world coordinates.} }
    \label{Fig:overview}
    \vspace{-9pt}
\end{figure}

\noindent
\textbf{Consistency evaluation.} Although we perform the augmentation on the training set, the predictions of the same containers should be consistent at the test time, which indicates a good generalization. Given the predictions of $n$-th video frame $t$ with container $k$, $\hat{Y}_{t,k}^n$, and its real-value label, $Y_k$, we measure the total variance, $Var$, as:

\begin{equation}
	{Var} = \sum\limits_{k}\sum\limits_{n}\sum\limits_{t}(\hat{Y}_{t,k}^n - {Y}_k)^2.
\end{equation}

\noindent
Using the estimated variance of the models on our manually chosen validation set, we select the model with high accuracy with low variance to ensure the generalization ability.

\vspace{-5pt}
\section{Experiments}
\label{sec:pagestyle}
\vspace{-5pt}
\subsection{Setup}
We evaluate our models on the CCM dataset, which is divided into the public train (9 containers, 684 configurations), public test (3 unseen containers, 228 configurations), and private test (3 unseen containers, 228 configurations). We then split the training set into two parts: 7 objects for training and 2 objects for self-validation. For object detection, we use YOLO-v5 pre-trained on COCO dataset ~\cite{lin2014microsoft}. As classes in COCO dataset are different from those in the CCM dataset, similar to ~\cite{iashin2021top}, we detect objects according to the categories: \textit{cups}, \textit{wine}, \textit{glasses}, \textit{vases} (for \textit{wine glasses}), \textit{books}, and \textit{laptops} (for \textit{boxes}). If object detection is failed, we use the average prediction from the training dataset. For training, we use Adam~\cite{kingma2014adam} optimizer with a learning rate of $0.001$ for 100 epochs for each task except the filling type classification trained for 200 epochs. We use cross-entropy loss for the filling level and type classification and $\ell 1$ loss the container capacity, mass, dimension estimation.

\subsection{Implementation Details}
For the filling level and type classification, we extract the log-Mel spectrogram of 64 dimensions from 8-channel audios. The fully-connected layers in the filling type classification contain two layers of size 512$\times$128 and 128$\times N$, where $N$ is the number of classes, and in the filling level classification the size becomes 1280$\times N$ with 0.2 dropout rate. The LSTM network contains 3 layers of 256 units each with 0.15 dropout rate. For the MobileNetv2 backbone, we use the original structure~\cite{sandler2018mobilenetv2}, except the last output layer that is modified to fit the desired output. The feature of the audio and color (RGB) image extracted by MobileNet is (1280,1) after average pooling. For the container capacity, mass and dimensions estimation, we use the depth information provided by the CCM dataset, i.e. the distance from the object to the camera. To handle the inaccurate depth information, e.g. the flattened depth of a cylindrical surface, we concatenate the depth map with the RGB channels and process them in the network in the same way.

\begin{table}[t]
\centering
\caption{Results of the filling type and level classification and container capacity estimation. The best results are highlighted in bold. Key-- A: audio, C: color image, D: depth image.}
\label{task123}
\resizebox{0.9\linewidth}{!}{%
\begin{tabular}{c cc cc cc}
\hline
\multirow{2}*{Method}     & \multicolumn{2}{c}{Filling Type}   & \multicolumn{2}{c}{Filling Level}  & \multicolumn{2}{c}{Capacity}       \\
\cmidrule(lr){2-3}
\cmidrule(lr){4-5}
\cmidrule(lr){6-7}
      & Input   & Score & Input   & Score & Input   & Score       \\
\hline
KEIO-ICS~\cite{Matsubara_GC_ICASSP22}      &A &80.72          &A& 65.73          &C& \textbf{72.26}        \\
BIT~\cite{iashin2021top}           &A& 94.26          &A,C& 79.65          &C& 60.57                 \\
HVRL~\cite{ishikawa2021audio} &A& 96.95         &A & 78.65         &C,D & 54.79                 \\
Concatenation\cite{liu2021va2mass} &A,C& 41.83         &A,C & 42.70         &C & 62.57             \\
ACC~\cite{Donaher2021EUSIPCO_ACC}          &A & 94.50         &A & \textbf{80.84} &-& -                 \\
NTNU~\cite{Christmann2020NTNU}          &A& 86.89         &- & -           &D & 67.30              \\
\hline
{Ours}          &A& \textbf{99.13} &A,C& 77.40          &C,D& 59.51          \\
\hline
\end{tabular}%
}
\vspace{-9pt}
\vspace{-5pt}
\end{table}

\begin{table}[t]
\centering
\caption{Results of the container mass and dimension, width (top and bottom) and height estimation. The best results are in bold. Key-- A: audio, C: Color image, D: depth image.}
\label{task45}
\resizebox{0.95\linewidth}{!}{%
\begin{tabular}{cccccc}
\hline
         &Input & Mass  & Width (top) & Width (bottom) & Height  \\ \hline
KEIO-ICS~\cite{Matsubara_GC_ICASSP22} & C & 40.19 & 69.09     & 59.74        & 70.07     \\
Visual~\cite{Apicella_GC_ICASSP22}   & C,D &49.64 & -       & -          & -     \\ \hline
{Ours} & C,D & \textbf{58.78} & \textbf{80.01}  & \textbf{76.09} & \textbf{74.33}  \\ \hline
\end{tabular}%
}
\vspace{-9pt}
\end{table}

\subsection{Results}

Table~\ref{task123} and~\ref{task45} show the results of the existing methods and ours for the filling type and level classification and container capacity, mass, and dimension estimation.
The benchmark scores\footnote{\href{http://corsmal.eecs.qmul.ac.uk/challenge.html}{http://corsmal.eecs.qmul.ac.uk/challenge.html}}, averages of the public and private test dataset, are based on 10 performance scores that quantify the accuracy of the estimations and consider the impact of the estimations on the quality of human-to-robot handover and delivery of the container by the robot in simulator~\cite{pang2021towards}.

As shown in Table \ref{task123}, our method outperforms other methods for the filling type classification. ACC~\cite{Donaher2021EUSIPCO_ACC} shows the best result in the filling level estimation, which uses multiple classifiers to first detect the action by the human and then the filling type. The audio data can generalize well on different containers for the filling type classification. However, in the filling level estimation, the model using audio-only may be distracted by some background noise.
We examine our multi-modality strategy for the filling level estimation. Table~\ref{fusion} show the results of the ablation study. Combination with the vision-based encoder improves the generalization of our model, which makes the total score higher than the audio-based model.
Compared with KEIO-ICS~\cite{Matsubara_GC_ICASSP22}, we obtain a higher score using the audio as input only, showing the effectiveness of our backbone model.

For the container capacity estimation, KEIO-ICS~\cite{Matsubara_GC_ICASSP22} shows the best result. While HVRL~\cite{ishikawa2021audio} and KEIO-ICS use a model specific to solve each task, we use the same backbone for the five estimation tasks, which can be more generalizable. In Table~\ref{task45}, our method outperforms others for estimating the container mass and dimensions, i.e., the width at the top and bottom, and height of the container.

\subsection{Analysis}
We perform ablation studies to analyze the effect of the modality fusion and the proposed data augmentation.

\begin{table}[t]
\centering
\caption{Effect of input modalities. Key-- A: audio, C: color image.}
\vspace{3pt}
\label{fusion}
\resizebox{0.45\linewidth}{!}{%
\begin{tabular}{ccc}
\hline
{Method} & Input   & {Filling Level} \\
\hline
\multirow{2}*{Ours} & A          & 75.53        \\
            & A,C        & 77.40         \\
\hline
\end{tabular}%
}
\vspace{-9pt}
\end{table}

\begin{figure}[t]
    \centering
    \includegraphics[width=0.85\columnwidth]{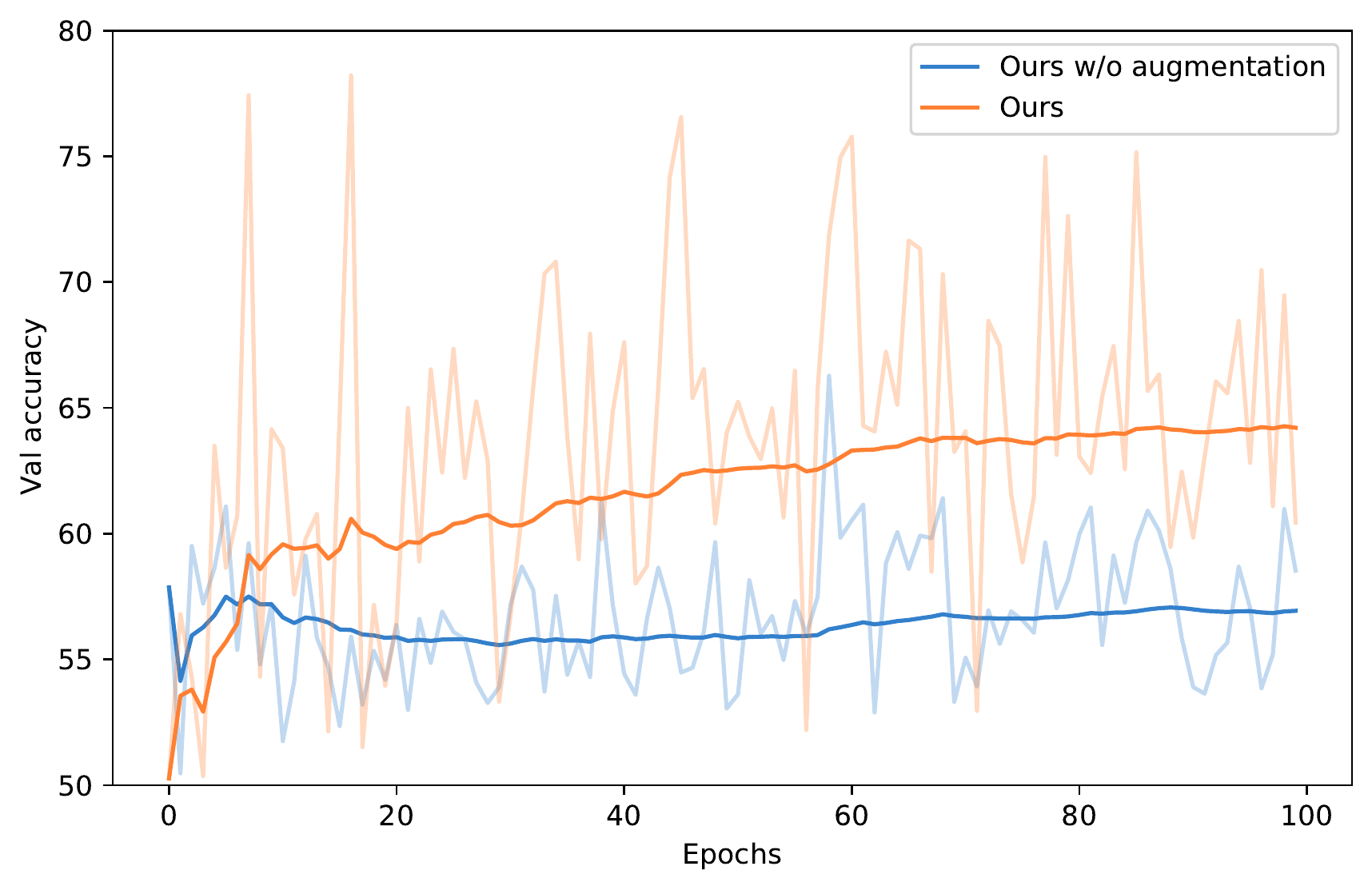}
    \vspace{-9pt}
    \caption{Effect of the proposed data augmentation on the capacity estimation. The train and validation set are manually selected with different containers using the public training set. The original curves (light color) are smoothed (dark color) for better visualization.}
    \label{augmentation}
    \vspace{-9pt}
\end{figure}

We examine the effect of the modality-fusion strategy in the filling level estimation in Table~\ref{fusion}. The results show that fusing the audio and visual features improves the performance of the filling level estimation as the vision cues can provide additional information, e.g., the filling level of the transparent containers.

We analyze the effect of the proposed data augmentation in our validation set (split from the training set). As shown in Figure~\ref{augmentation}, in the container capacity estimation, the validation accuracy of the model with the data augmentation is higher as the epochs increase compared to the model without the augmentation.

\section{Conclusion}
\vspace{-5pt}
We presented methods to estimate the physical properties of the household containers for human-robot collaboration. The key idea of our approach is to improve the generalization ability of the models using multi-modal information with attention and geometric relationship between 2D images and 3D containers. Future works include the investigation of a physically-correct data augmentation method and transfer learning on each task.

\vspace{9pt}
\noindent
\textbf{Acknowledgement.} This project made use of time on Tier 2 HPC facility JADE2, funded by EPSRC (EP/T022205/1).

\vfill\pagebreak


\begin{thebibliography}{10}

\bibitem{sanchez2020benchmark}
R.~Sanchez-Matilla, K.~Chatzilygeroudis, A.~Modas, N.F. Duarte, A.~Xompero,
  P.~Frossard, A.~Billard, and A.~Cavallaro,
\newblock ``Benchmark for human-to-robot handovers of unseen containers with
  unknown filling,''
\newblock {\em IEEE Robot. Autom. Lett.}, vol. 5, no. 2, pp. 1642--1649, 2020.

\bibitem{medina2016human}
J.R. Medina, F.~Duvallet, M.~Karnam, and A.~Billard,
\newblock ``A human-inspired controller for fluid human-robot handovers,''
\newblock in {\em IEEE-RAS Int. Conf. Humanoid Robot.}, 2016.

\bibitem{rosenberger2020object}
P.~Rosenberger, A.~Cosgun, R.~Newbury, J.~Kwan, V.~Ortenzi, P.~Corke, and
  M.~Grafinger,
\newblock ``Object-independent human-to-robot handovers using real time robotic
  vision,''
\newblock {\em IEEE Robot. Autom. Lett.}, vol. 6, no. 1, pp. 17--23, 2020.

\bibitem{ortenzi2021object}
V.~Ortenzi, A.~Cosgun, T.~Pardi, W.P. Chan, E.~Croft, and D.~Kuli{\'c},
\newblock ``Object handovers: a review for robotics,''
\newblock {\em IEEE Trans. Robot.}, 2021.

\bibitem{yang2020reactive}
W.~Yang, C.~Paxton, A.~Mousavian, Y.~Chao, M.~Cakmak, and D.~Fox,
\newblock ``Reactive human-to-robot handovers of arbitrary objects,''
\newblock in {\em IEEE Int. Conf. Robot. Autom.}, 2021.

\bibitem{pang2021towards}
Y.L. Pang, A.~Xompero, C.~Oh, and A.~Cavallaro,
\newblock ``Towards safe human-to-robot handovers of unknown containers,''
\newblock in {\em IEEE Int. Conf. Robot Hum. Interact. Commun.}, 2021.

\bibitem{liu2021va2mass}
Q.~Liu, F.~Feng, C.~Lan, and R.H. Chan,
\newblock ``Va2mass: Towards the fluid filling mass estimation via integration
  of vision and audio learning,''
\newblock in {\em Int. Conf. Pattern Recognit.}, 2021.

\bibitem{ishikawa2021audio}
R.~Ishikawa, Y.~Nagao, R.~Hachiuma, and H.~Saito,
\newblock ``Audio-visual hybrid approach for filling mass estimation,''
\newblock in {\em IEEE Conf. Comput. Vis. Pattern Recognit.}, 2021.

\bibitem{iashin2021top}
V.~Iashin, F.~Palermo, G.~Solak, and C.~Coppola,
\newblock ``Top-1 corsmal challenge 2020 submission: Filling mass estimation
  using multi-modal observations of human-robot handovers,''
\newblock in {\em Int. Conf. Pattern Recognit. Workshop}, 2021.

\bibitem{corsmal_dataset}
A.~Xompero, R.~Sanchez-Matilla, R.~Mazzon, and A.~Cavallaro,
\newblock ``Corsmal containers manipulation (1.0) [dataset],''
\newblock {\em DOI: https://doi.org/10.17636/101CORSMAL1}, 2020.

\bibitem{sandler2018mobilenetv2}
M.~Sandler, A.~Howard, M.~Zhu, A.~Zhmoginov, and L.C. Chen,
\newblock ``Mobilenetv2: Inverted residuals and linear bottlenecks,''
\newblock in {\em IEEE Conf. Comput. Vis. Pattern Recognit.}, 2018.

\bibitem{hou2021coordinate}
Q.~Hou, D.~Zhou, and J.~Feng,
\newblock ``Coordinate attention for efficient mobile network design,''
\newblock in {\em IEEE/CVF Conf. Comput. Vis. Pattern Recognit.}, 2021.

\bibitem{logan2000mel}
B.~Logan,
\newblock ``Mel frequency cepstral coefficients for music modeling,''
\newblock in {\em Int. Symp. Music Inf. Retr.}, 2000.

\bibitem{yolov5}
G.~Jocher, K.~Nishimura, T.~Mineeva, and R.~Vilari{\~n}o,
\newblock ``Yolov5,''
\newblock {\em Code repository https://github. com/ultralytics/yolov5}, 2020.

\bibitem{hochreiter1997long}
S.~Hochreiter and J.~Schmidhuber,
\newblock ``Long short-term memory,''
\newblock {\em Neural Comput.}, vol. 9, no. 8, pp. 1735--1780, 1997.

\bibitem{vgg}
K.~Simonyan and A.~Zisserman,
\newblock ``Very deep convolutional networks for large-scale image
  recognition,''
\newblock in {\em Int. Conf. Learn. Represent.}, 2015.

\bibitem{hu2018squeeze}
J.~Hu, L.~Shen, and G.~Sun,
\newblock ``Squeeze-and-excitation networks,''
\newblock in {\em IEEE Conf. Comput. Vis. Pattern Recognit.}, 2018.

\bibitem{deng2009imagenet}
J.~Deng, W.~Dong, R.~Socher, L.J. Li, K.~Li, and L.~Fei-Fei,
\newblock ``Imagenet: A large-scale hierarchical image database,''
\newblock in {\em IEEE Conf. Comput. Vis. Pattern Recognit.}, 2009.

\bibitem{lin2014microsoft}
T.~Lin, M.~Maire, S.~Belongie, J.~Hays, P.~Perona, D.~Ramanan, P.~Dollr, and
  C.L. Zitnick,
\newblock ``Microsoft coco: Common objects in context,''
\newblock in {\em Eur. Conf. Comput. Vis.}, 2014.

\bibitem{kingma2014adam}
D.P. Kingma and J.~Ba,
\newblock ``Adam: A method for stochastic optimization,''
\newblock {\em arXiv preprint arXiv:1412.6980}, 2014.

\bibitem{Matsubara_GC_ICASSP22}
T.~Matsubara, S.~Otsuki, Y.~Wada, H.~Matsuo, T.~Komatsu, Y.~Iioka, K.~Sugiura,
  and H.~Saito,
\newblock ``{Shared transformer encoder with mask-based 3D model estimation for
  container mass estimation},''
\newblock in {\em \textit{IEEE Int. Conf. Acoustic, Speech, Sig. Proc. Grand
  Challenges: Audio-Visual Object Classification For Human-Robot
  Collaboration}}, 2022.

\bibitem{Donaher2021EUSIPCO_ACC}
S.~Donaher, A.~Xompero, and A.~Cavallaro,
\newblock ``Audio classification of the content of food containers and drinking
  glasses,''
\newblock in {\em Europ. Signal Proc. Conf.}, 2021.

\bibitem{Christmann2020NTNU}
G.~Christmann and J.~Song,
\newblock ``{2020 CORSMAL Challenge - Team NTNU-ERCReport},'' 2020,
\newblock
  \url{https://corsmal.eecs.qmul.ac.uk/resources/challenge/2020.11.30_CORSMAL_NTNU-ERC_Report.pdf}.

\bibitem{Apicella_GC_ICASSP22}
T.~Apicella, G.~Slavic, R.~Ragusa, P.~Gastaldo, and L.~Marcenaro,
\newblock ``{Container localisation and mass estimation from an RGB-D
  camera},''
\newblock in {\em \textit{IEEE Int. Conf. Acoustic, Speech, Sig. Proc. Grand
  Challenges: Audio-Visual Object Classification For Human-Robot
  Collaboration}}, 2022.

\end{thebibliography}
\end{document}